\documentclass[letterpaper]{article} 
\usepackage{aaai23}
\usepackage{times}  
\usepackage{helvet}  
\usepackage{courier}  
\usepackage[hyphens]{url}  
\usepackage{graphicx} 
\usepackage{natbib}  
\usepackage{caption} 
\usepackage{algorithm}
\usepackage{algorithmic}
\usepackage{cite}

\usepackage{microtype}

\usepackage[table,dvipsnames]{xcolor}
\usepackage{pgfplots}
\usepackage{dsfont}
\usepackage{url}
\usepackage{booktabs} 
\usepackage{graphicx}
\usepackage{tabularx}
\usepackage{amsmath}
\usepackage{bbm}
\usepackage{soul}
\usepackage{color, xcolor}
\usepackage{multirow, makecell, caption}
\usepackage{arydshln}
\usepackage{longtable}
\usepackage{booktabs}

\usepackage[normalem]{ulem}
\usepackage{amssymb}
\usepackage{amsfonts}
\usepackage{multicol}
\usepackage{pifont}

\usepackage{csquotes}
\usepackage{xspace}
\usepackage{paralist}
\usepackage{mdwlist}
\usepackage{subfigure}
\usepackage{makecell}
\usepackage{array}
\usepackage{enumitem}
\usepackage{bm}
\usepackage{graphicx}

\usepackage{cleveref}

\usepackage{newfloat}
\usepackage{listings}
\DeclareCaptionStyle{ruled}{labelfont=normalfont,labelsep=colon,strut=off} 
\floatstyle{ruled}
\newfloat{listing}{tb}{lst}{}
\floatname{listing}{Listing}
\newcommand{\xmark}{\ding{55}}
\newcommand{\cmark}{\ding{51}}

\newcommand{\ie}{\textit{i.e.}}
\newcommand{\eg}{\textit{e.g.}}
\newcommand{\method}{\textsc{VenCE}\xspace}
\newcommand{\dataset}{\textsc{FecData}\xspace}

\makeatletter
\def\adl@drawiv#1#2#3{%
        \hskip.5\tabcolsep
        \xleaders#3{#2.5\@tempdimb #1{1}#2.5\@tempdimb}%
                #2\z@ plus1fil minus1fil\relax
        \hskip.5\tabcolsep}
\newcommand{\cdashlinelr}[1]{%
  \noalign{\vskip\aboverulesep
           \global\let\@dashdrawstore\adl@draw
           \global\let\adl@draw\adl@drawiv}
  \cdashline{#1}
  \noalign{\global\let\adl@draw\@dashdrawstore
           \vskip\belowrulesep}}
\makeatother

\title{
\textit{Converge to the Truth}:\\
Factual Error Correction via Iterative Constrained Editing
}

\author {
    Jiangjie Chen\textsuperscript{\rm 1}\thanks{Authors contributed equally.}\thanks{Work done during internship at ByteDance AI Lab.},
    Rui Xu\textsuperscript{\rm 1}\footnotemark[1], 
    Wenxuan Zeng\textsuperscript{\rm 2},
    Changzhi Sun\textsuperscript{\rm 3}\thanks{Corresponding authors.},
    Lei Li\textsuperscript{\rm 4},
    Yanghua Xiao\textsuperscript{\rm 1,5}\footnotemark[3]
}
\affiliations {
    \textsuperscript{\rm 1}Shanghai Key Laboratory of Data Science, School of Computer Science, Fudan University\\
    \textsuperscript{\rm 2}University of Electronic Science and Technology of China\\
    \textsuperscript{\rm 3}ByteDance AI Lab \\
    \textsuperscript{\rm 4}University of California, Santa Barbara\\
    \textsuperscript{\rm 5}Fudan-Aishu Cognitive Intelligence Joint Research Center\\
    jjchen19@fudan.edu.cn, ruixu21@m.fudan.edu.cn, wenxuan.zeng@std.uestc.edu.cn, \\
    sunchangzhi@bytedance.com, leili@cs.ucsb.edu, shawyh@fudan.edu.cn\\
}

\usepackage{bibentry}

\begin{document}

\maketitle

\begin{abstract}

Given a possibly false claim sentence, how can we automatically correct it with minimal editing?
Existing methods either require a large number of pairs of false and corrected claims for supervised training or do not handle well errors spanning over multiple tokens within an utterance. 
In this paper, we propose \method, a novel method for factual error correction (FEC) with minimal edits. 
\method formulates the FEC problem as iterative sampling editing actions with respect to a target density function. 
We carefully design the target function with predicted truthfulness scores from an offline trained fact verification model. 
\method samples the most probable editing positions based on back-calculated gradients of the truthfulness score concerning input tokens and the editing actions using a distantly-supervised language model (T5). 
Experiments on a public dataset show that \method improves the well-adopted SARI metrics by 5.3 (or a relative improvement of 11.8\%) over the previous best distantly-supervised methods.
Resources of \method can be found at \url{https://github.com/jiangjiechen/VENCE}.
\end{abstract}

\section{Introduction}
\label{sec:intro}

\begin{figure}[t]
    \centering
    \includegraphics[width=\linewidth]{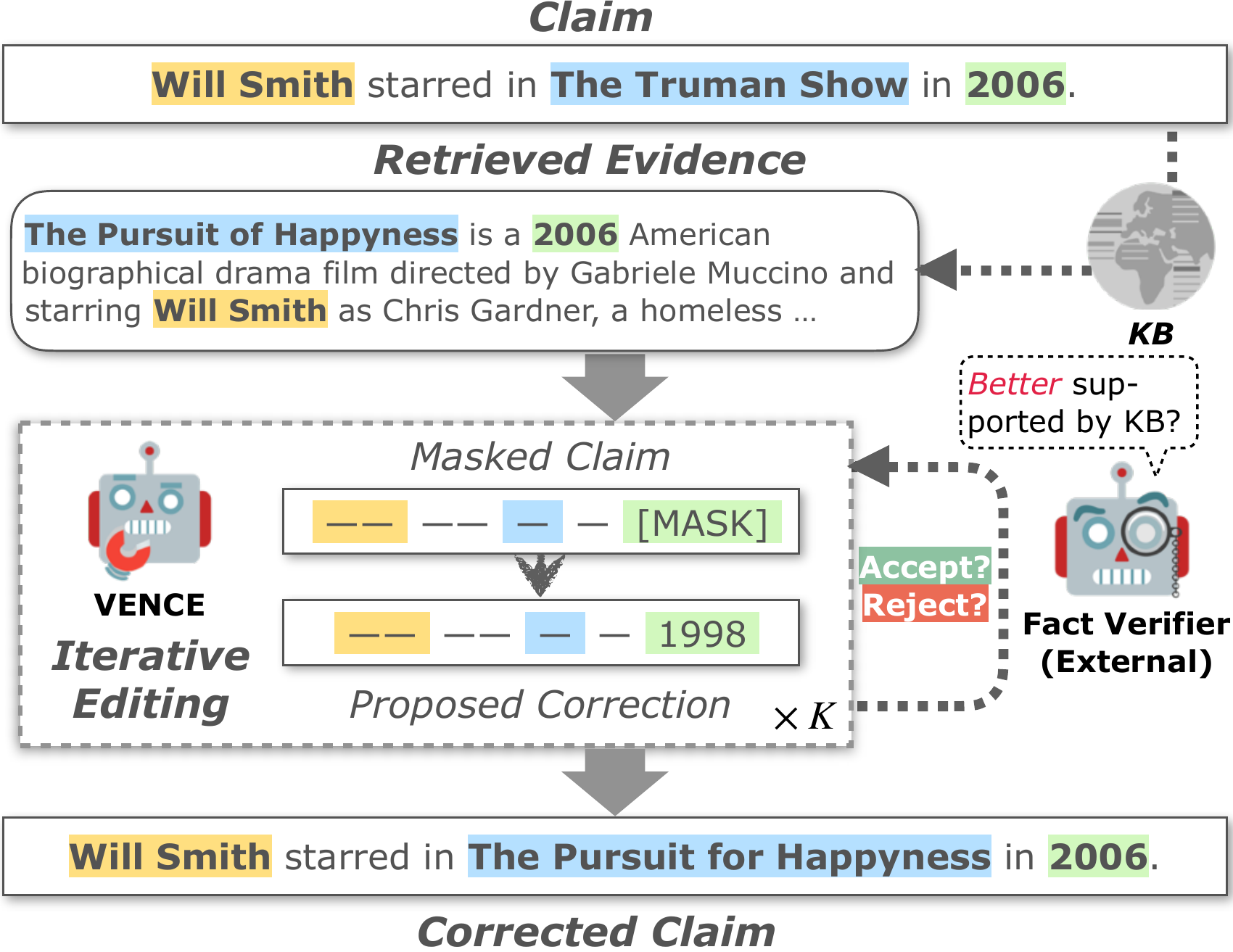}
    \caption{An overview of \method, an FEC method based on Metropolis-Hastings sampling. 
    Given a claim and retrieved evidence texts, \method iteratively corrects the claim by claim masking and correction proposal.
    Based on whether a new proposal is better supported by the evidence, an external fact verifier contributes to the decision of proposal acceptance.
    After the sampling is converged, the most supported proposal is selected as the corrected claim.}
    \label{fig:front}
\end{figure}


In the digital age, misinformation with textual \textit{factual errors} is spreading widely and quickly with the help of social media.
However, current text correction approaches mostly focus on \textit{grammatical error correction} (GEC)~\cite{yuan-briscoe-2016-grammatical,Chollampatt_Ng_2018,omelianchuk-etal-2020-gector,yasunaga-etal-2021-lm}.
The automatic correction of \textit{factual errors} may help prevent them from fueling misunderstanding and misleading decision-making, attracting an increasing amount of research attention.
Given a false claim and its evidence texts from trustworthy knowledge bases, \textit{factual error correction} (FEC) aims to correct the nonfactual text spans of the claim, so that it is better supported by evidence.
For the example in Figure~\ref{fig:front}, an FEC system needs to correct that the movie that Will Smith starred in in 2006 is \textit{The Pursuit for Happyness}, not \textit{The Truman Show}.

The main factor limiting the application of FEC systems is the cost to obtain high-quality fine-grained annotations, making it relatively under-explored.
To gain more training data, instead of annotating claims paired with their corrections, researchers~\cite{Shah_Schuster_Barzilay_2020,thorne-vlachos-2021-evidence} start using distant supervision from fact verification (FV) datasets~\cite{thorne-etal-2018-fever}.
Specifically, for a claim \textsc{Supported} by evidence in the FV dataset, they first mask some tokens in the claim and train the FEC system to restore the original claim from the masked one.

Existing distantly supervised models usually follow a one-pass mask-then-correct paradigm \cite{Shah_Schuster_Barzilay_2020,thorne-vlachos-2021-evidence}.
The masker is designed to find problematic spans in the claim, where token-level explanations of the FV model are usually exploited as masks.
The corrector generates the corrected sentence given the masked claim.
However, those models are limited due to two reasons:
\begin{inparaenum}[\it 1)]
    \item The problem of semantic drift, \ie, over-correcting the claim and making the intended meaning of the claim broken.
    \item As a response to the lack of annotated data for FEC, the supervision from fact verification is under-utilized.
\end{inparaenum}


To address these problems, we propose to integrate the advantages of both \textit{fact verification} and \textit{iterative text editing} for FEC.
\textit{Our first key insight} is to correct errors in the manner of iterative editing.
We break the correction process into unit-level (\eg, tokens and entities), to overcome the over-erasure problem and explore and revise more correction choices.
\textit{Our second key insight} is to bridge FV with FEC in the iterative editing framework, where FV offers control and guidance to the correction in each editing iteration.
Since resources for the FV task are significantly richer than that of FEC, an FEC system can then be able to iteratively edit small errors under the constraint of fact verification, guiding towards correctness during each iteration.

Motivated by these ideas, in this paper, we propose \method, a \textbf{VE}rificatio\textbf{N}-guided \textbf{C}onstrained \textbf{E}diting system for factual error correction.
To realize our ideas, an FEC model should be able to utilize external constraints (FV) without direct supervision from FEC.
Thus, we build our system based on the framework of constrained text generation with Metropolis-Hastings (MH) sampling~\cite{metropolis1953equation}, a typical Markov chain Monte Carlo method, which proved to be effective on various unsupervised constrained text generation tasks~\cite{Miao_Zhou_Mou_Yan_Li_2019,zhang-etal-2020-language-generation,Chen_Gan_Cheng_Zhou_Xiao_Li_2022,mireshghallah-etal-2022-mix}.
Compared with previous work, \method utilizes an external FV model for guidance in two ways:
\begin{inparaenum}[\it 1)]
    \item the token gradients from the FV model guide the finding of edit tokens, and
    \item the truthfulness score by the FV model contributes to a probabilistic energy-based model~\cite{hinton2002training} that decides the acceptance of a correction proposal in each iteration.
\end{inparaenum}
To maintain the fluency and minimum modification to the original claim (as opposed to over-erasure), we also consider a language modeling score and hamming distance in the energy model, respectively.
To overcome the challenge of editing multi-token entities, we separate token space with entity space, and train a multi-tasking T5 model \cite{raffel-etal-2020-exploring} to propose from distributions of both spaces.

To summarize, this work contributes to solving factual error correction without direct supervision.
To our knowledge, we are the first to adopt an iterative text editing method (\method) for this task, which alleviates the over-erasure problem in previous methods.
Also, \method enjoys a more powerful error revision ability by effectively integrating external but coarse-grained verification signals during each editing iteration.
Experimental results demonstrate the effectiveness of \method over previous methods, which achieves a new state-of-the-art for distantly supervised FEC task.

\section{Related Work}
\label{sec:related}

\paragraph{Text Factuality}
The factuality of text has been widely studied by the community.
During the automatic production of text, text generation systems are prone to hallucinate, harming the factuality of generated text~\cite{raunak-etal-2021-curious,xiao-wang-2021-hallucination}. 
Methods have been proposed to alleviate hallucination (to the internal or external knowledge) in generation for text summarization~\cite{Cao_Wei_Li_Li_2018,cao-etal-2022-hallucinated}, dialogue systems~\cite{honovich-etal-2021-q2}, text simplification~\cite{devaraj-etal-2022-evaluating}, etc.
In the post-verification stage, fact verification (or fact checking) aims to evaluate factuality of existing texts, which leverages external evidence to verify a given claim~\cite{thorne-etal-2018-fever,augenstein-etal-2019-multifc,wadden-etal-2020-fact,aly-etal-2021-fact}.
It is worth noting that, many recent studies focus on interpretable fact verification~\cite{stammbach2020fever,kotonya-toni-2020-explainable,samarinas-etal-2021-improving,wu-etal-2021-unified}, which is related to the objective of FEC - understanding and making the claims better supported by evidence.
For example, ProoFVer~\cite{krishna-riedel-vlachos-2021-proofver} is trained to generate intermediate natural logic proofs for verification, where the proofs are annotated,
while \textsc{LoReN}~\cite{Chen_Bao_Sun_Zhang_Chen_Zhou_Xiao_Li_2022} decomposes claim verification at phrase-level by logic and learns phrase veracity predictions without any intermediate annotations.
Also aiming for the text factuality problem, we tackle it from the perspective of post-correction, which verifies and corrects nonfactual texts.

\paragraph{Textual Error Correction}
According to error types, research on textual error correction can be categorized as grammatical error correction (GEC) and factual error correction (FEC).
GEC~\cite{ng-etal-2014-conll,omelianchuk-etal-2020-gector,ishii-etal-2021-grammatical,qorib-etal-2022-frustratingly} aims to correct grammatical errors without changing the semantic content of each sentence. 
Existing literature generally relies on human-labeled data, while \citet{yasunaga-etal-2021-lm} study GEC under an unsupervised paradigm, which leverages a critic model to assess the input and learns to repair bad examples without labeled data. 
Unlike GEC, FEC needs to modify the semantic content of sentences to achieve factual consistency against trustworthy knowledge sources, either textual corpus~\cite{Shah_Schuster_Barzilay_2020,thorne-vlachos-2021-evidence} or structured tables~\cite{hayate-qiao-li-2020-fact}. 
In this work, we study the problem of FEC based on textual knowledge bases.

\paragraph{Text Editing}

Many research efforts have been made to text editing~\cite{kasner-dusek-2020-data,higashiyama-etal-2021-text,agrawal-carpuat-2022-imitation}, especially for tasks that source and target text that has high overlap,
such as sentence simplification~\cite{inui-etal-2003-text,mallinson2022edit5}, translation post-editing~\cite{NEURIPS2019_675f9820,lee-etal-2020-noising}, story rewriting~\cite{Chen_Gan_Cheng_Zhou_Xiao_Li_2022}, text style transfer~\cite{hu-etal-2022-text}, etc.
To impose constraints during editing, in recent years, a prevailing way is to adopt Markov chain Monte Carlo sampling for text generation.
\citet{Miao_Zhou_Mou_Yan_Li_2019} first utilize Metropolis-Hastings (MH) sampling for constrained text generation, which edits discrete tokens under the desired stationary distributions as constraints.
As an extension, \citet{zhang-etal-2020-language-generation} manage to satisfy combinatorial constraints for text generation, and \citet{mireshghallah-etal-2022-mix} propose a global score-based model that also leverages MH sampling for controllable text generation.
We follow this line of work and extend with verification signals as constraints and guidance for FEC, and extends token-level editing to entity-level for semantic consistency.

\section{Method}
\label{sec:method}

\begin{figure}[t]
    \centering
    \includegraphics[width=\linewidth]{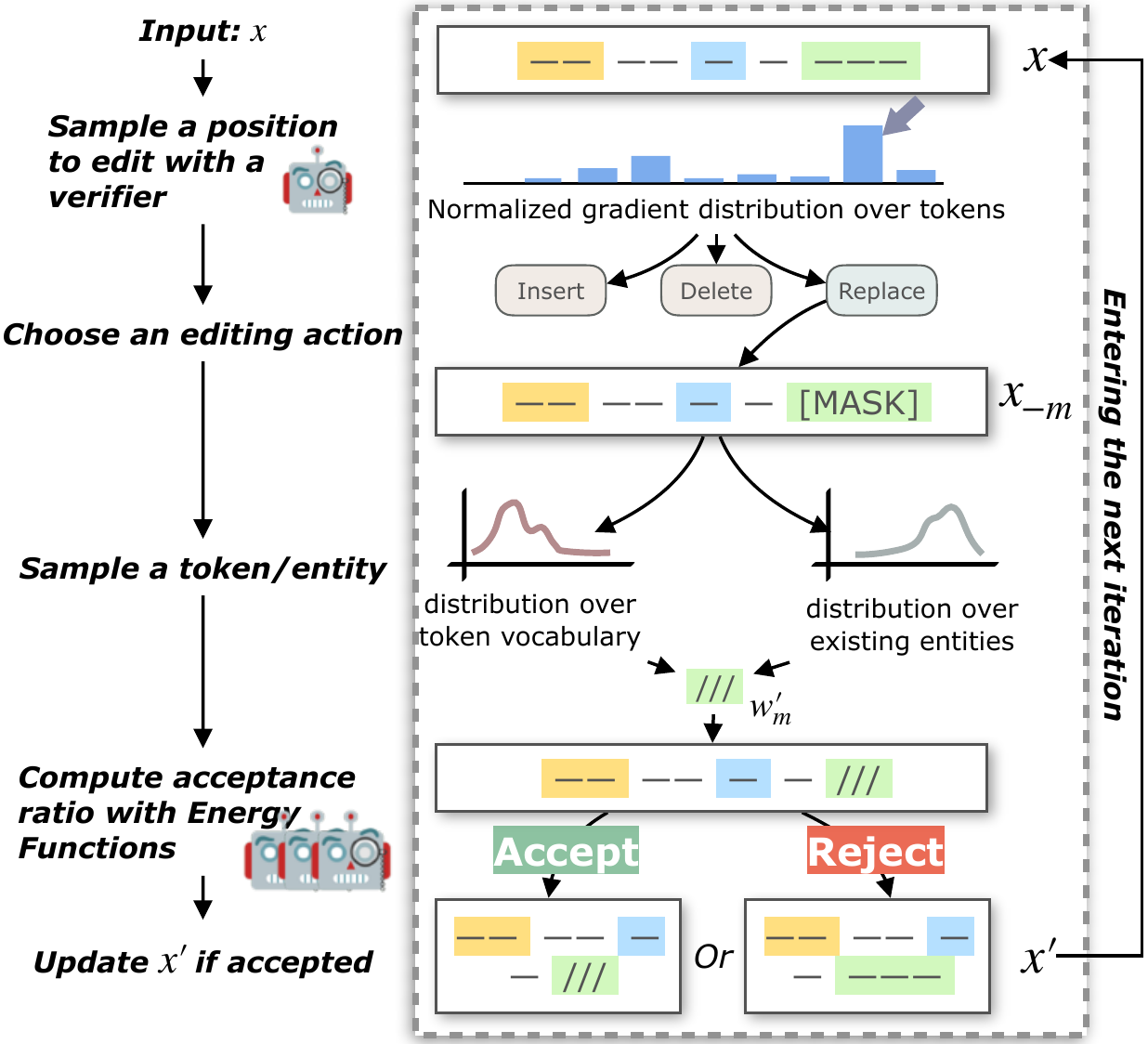}
    \caption{The editing procedure of \method of each iteration.}
    \label{fig:method}
\end{figure}

In this section, we present our FEC method \method.
We will first sketch an overview in $\mathsection$~\ref{sec:overview}. 
Then, we will detail the implementations in the sketch in $\mathsection$~\ref{sec:energy}-\ref{sec:corrector}.

\subsection{Overview}
\label{sec:overview}

The corrected sentence should be fluent, truthful and minimally edited, which are the constraints to be satisfied during correction.
To this end, we adopt a text editing framework based on Metropolis-Hastings sampling~\cite{metropolis1953equation,Miao_Zhou_Mou_Yan_Li_2019}, a MCMC method.
The MH algorithm is able to ensure the edited text to be converged where the above constraints are satisfied, through the custom definition of the \textit{stationary distribution} of the Markov chain.

Given an original claim $x^0$ and retrieved evidence texts $E$, we iteratively execute steps of \textit{mask-then-correct} over the claim.
The correction proposal for the claim ($x = \{w_1, \cdots, w_{|x|}\}$) in current iteration has the following steps:
\begin{enumerate}
    \item Sample an edit position $m$ to be masked within $x$, \ie, $P_1(m\mid x), m = 1, 2, ..., |x|$.
     This yields a masked claim $x_{-m} = \{w_1, \cdots, w_{m-1}, \texttt{[MASK]}, \cdots, w_{|x|}\}$ ($\mathsection$~\ref{sec:masker});
    \item Sample an edit action $a$ (\textit{insert}, \textit{delete} or \textit{replace}), \ie, $P_2(a)$ ($\mathsection$~\ref{sec:corrector});
    \item Sample a corrected token/entity according to the masked claim $x_{-m}$ and the sampled edit action $a$,  and get $x^\prime$, \ie, $P_3(x^\prime\mid x_{-m}, a)$ ($\mathsection$~\ref{sec:corrector}).
\end{enumerate}
Together, we have the \textit{transition distribution} $g(x^\prime \mid x)$ in the Markov chain taking the action $a$ to edit position $m$:
\begin{equation}
    g(x^\prime\mid x) = P_1(m\mid x) P_2(a)P_3(x^\prime\mid x_{-m}, a).
\end{equation}
Note that we omit evidence $E$ in the condition for brevity.

In order to make sure the sampling to be converged on the \textit{stationary distribution} $\pi(x)$, the acceptance of the proposed $x^\prime$ should be decided by the \textit{acceptance ratio} $A(x^\prime \mid x)$, and the accepted $x^\prime$ will enter the next iteration.
We implement $\pi(x)$ with the target objectives of FEC (factual, fluent and minimally modified), and use a probabilistic energy-based model $\mathcal{E}(x)$, where $\pi(x) = \frac{e^{-\mathcal{E}(x)}}{Z}$~\cite{mireshghallah-etal-2022-mix}, the Boltzmann distribution of the energy functions ($\mathsection$~\ref{sec:energy}).
Thus, $A(x^\prime \mid x)$ is defined as in the MH algorithm:
\begin{equation}
    \begin{split}
        A(x^\prime \mid x)&=\min \{1, \frac{\pi(x^\prime) g(x \mid x^\prime)}{\pi(x) g(x^\prime \mid x)}\} \\
        &=\min \{1, \frac{e^{-\mathcal{E}(x^\prime)} g(x \mid x^\prime)}{e^{-\mathcal{E}(x)} g(x^\prime \mid x)}\}.
    \end{split}
\end{equation}

Finally, the accepted texts are ranked based on $\mathcal{E}(x)$, and the highest one is chosen as the output.
We visualize the process within each iteration in Figure \ref{fig:method}.
This process goes on until converged (20 iterations in practice).
Keeping this workflow in mind, next, we will place the pieces together.

\subsection{Energy Functions for Target Objectives}
\label{sec:energy}

First and foremost, we start by defining the target objectives in FEC.
A corrected claim $x$ should be fluent ($\mathcal{E}_\mathtt{LM}(x)$), truthful according to trustworthy evidence ($\mathcal{E}_\mathtt{V}(x)$) and edited with minimum modifications to preserve the intended meaning ($\mathcal{E}_\mathtt{H}(x)$), therefore we have an aggregated energy function $\mathcal{E}(x)$ defined as:
\begin{equation}
    \mathcal{E}(x) = \mathcal{E}_\mathtt{LM}(x) + \mathcal{E}_\mathtt{V}(x) + \mathcal{E}_\mathtt{H}(x).
\end{equation}

\paragraph{Language Modeling Score}

Constrained text generation with MH sampling should first guarantee text fluency, as shown by various previous work~\cite{Miao_Zhou_Mou_Yan_Li_2019,Chen_Gan_Cheng_Zhou_Xiao_Li_2022,mireshghallah-etal-2022-mix}.
Following \citet{goyal2022exposing,mireshghallah-etal-2022-mix}, we implement the energy function as
\begin{equation}
    \mathcal{E}_\mathtt{LM}(x)= - \sum_i \log P_\mathtt{MLM}(w_i\mid x_{-i})    
\end{equation}
using a masked language model (MLM) with a BERT-base \cite{devlin-etal-2019-bert}, which is the sum of the predicted logits over every masked token $w_i$.
Note that this score can also be implemented with a causal language model such as GPT-2 \cite{radford2019language}.

\paragraph{Truthfulness Score}

An important point of view of our method is to incorporate verification signals deeply into each edit iteration.
Therefore, the truthfulness score is indispensable for the energy function.
To achieve this goal, we use an off-the-shelf discriminative verification model $P_\mathtt{V}$ that predicts whether the current claim $x$ is supported by the evidence $E$, yielding the probability of the \textsc{Supported} label.
Thus, the energy function is defined as:
\begin{equation}
    \mathcal{E}_\mathtt{V}(x) = - \log P_\mathtt{V}(\textsc{Supported}\mid x, E).
\end{equation}
In this way, powerful verifiers can be exploited to control the factuality of each correction proposal, since resources for verification datasets are generally much richer than that of the FEC task.

\paragraph{Hamming Distance}

One tricky pitfall for this task is that the corrections could be too often to keep the intended meaning of the original claim.
For example, one can replace with a brand new correct text but completely different than the previous claim.
Therefore, we also aim to constrain the number of edits of the proposed text, where we calculate a hamming distance between current claim $x$ and the original claim $x^0$, 
\begin{equation}
    \mathcal{E}_\mathtt{H}(x) = \textsc{HammingDistance}(x, x^0).
\end{equation}

\subsection{Editing Position Proposal}
\label{sec:masker}

In order to implement the transition distribution $g(x^\prime\mid x)$, we start by designing $P_1(m\mid x)$, where only a handful of masks are applied during each iteration.

To find the position to edit, the MH algorithm generally randomly samples from the token sequence and edits a token at each time.
However, this yields two challenges for the FEC task:
\begin{inparaenum}[\it 1)]
    \item most tokens do not need to be changed: \textit{how to accelerate the sampling process with only necessary edits?} and
    \item factual errors usually exist in the wrong multi-token entities: \textit{how to edit consecutive tokens in an entity in MH sampling?}
\end{inparaenum}

\paragraph{Sampling from Gradient-based Distribution}

To address the first challenge, we assume that the gradient by the verification model $P_\mathtt{V}$ of erroneous tokens should be larger than the correct ones, inspired by \citet{sha-2020-gradient}.
Therefore, we reuse the $\mathcal{E}_\mathtt{V}$ from $\mathsection$~\ref{sec:energy} and substitute the random sampling with a normalized distribution over tokens of the token gradient calculated by back-propagating $P_\mathtt{V}$.
Then we have 
\begin{equation}
    P_1(i\mid x) \propto {\Vert \nabla_{\bm{w}_i} \mathcal{E_\mathtt{V}}(x) \Vert_2}
\end{equation}
where $\bm{w}_i$ is the word representation of the token $w_i$ and $\Vert \cdot \Vert_2$ denotes vector 2-norm operation.

\paragraph{Multi-token Entity Masking}

As for the second challenge, we believe tokens in an entity should be edited together during each proposal.
Yet, sampling with one token at a time breaks the semantic meaning of the entity.
To solve this problem, we choose to mask all the tokens in each named entity at a time.
To do so, we first identify the named entities in the claim using an off-the-shelf NER model.~\footnote{\url{https://huggingface.co/Jean-Baptiste/camembert-ner}}
When any token within an entity is sampled, we select the whole entity for correction, \ie, ``Will Smith starred in \underline{\texttt{[MASK]}} in 2006'', other than ``Will Smith starred in \underline{The \texttt{[MASK]} show} in 2006''.
Thus, the probability of an entity $e$ to be masked is the sum of the probability of its tokens, \ie, $\sum_{w_i \in e} P_1(i\mid x)$.
The burden of correcting the multi-token entity is now shifted to the edit proposal step ($\mathsection$~\ref{sec:corrector}), which we will soon discuss.

\subsection{Editing Proposal}
\label{sec:corrector}
After getting the masked sequence $x_{-m}$ in each iteration, we will \textit{randomly} sample an action from the uniform distribution $P_2(a)$ of \textit{insertion}, \textit{deletion} and \textit{replacement}, and propose a token/entity from $P_3(x^\prime\mid x_{-m}, a)$.

However, with the aforementioned multi-token masking, challenges arise during sampling.
The Markov chain must satisfy \textit{detailed balance condition} in order for the sampling to converge on the stationary distribution $\pi(x)$ with the target objectives ($\mathsection$~\ref{sec:energy}).
To this end, the transition distribution $g(x^\prime\mid x)$ must be \textit{reversible}.
And yet, predicting an entity as a list of tokens given one mask is not reversible.
For example, the state with a broken entity (such as ``Will Smith starred in \textit{The Pursuit for}'') will never be reached during a multi-token insertion, but it is necessary during a token-by-token deletion.
Therefore, we separate entity space and token space with a generative proposal model, which we will next introduce.
The correctness of our approach will be clarified in $\mathsection$~\ref{sec:converge}.

\paragraph{Generative Proposal Model: Token vs. Entity}

Traditional proposal models (MLMs) have difficulty in dealing with multi-token prediction of a masked sentence.
Instead of widely-used masked language models for token proposal, we fine-tune a sequence-to-sequence generative language model (T5, \citealp{raffel-etal-2020-exploring}) to propose multi-token corrections.
We separate the sampling spaces for entities and tokens, and define two generation tasks: token generation and entity generation, which are distinguished by adding different prefixes, a property enjoyed by T5.~\footnote{Example prefixes: ``\textit{substituted word:}'' and ``\textit{substituted entity:}''.}
During the training of T5, we use distant supervision by \textsc{Supported} claims.
The model takes as input a masked sentence of a  and its evidence, and outputs either a token or entity substituting that mask.
The output choice depends on the types of correction action, which we will explain shortly.

With the proposal model, the transition distribution of corrected token/entity (denoted as $w^\prime_m$, and $x^\prime = w^\prime_m + x_{-m}$) is calculated accordingly.
For the single token proposal, the probability distribution $P^\mathrm{tok}_3(w^\prime_m\mid x_{-m}, a)$ naturally equals the vocabulary distribution of the token to be generated.
For entity proposal, however, the probability distribution $P^\mathrm{ent}_3(w^\prime_m\mid x_{-m}, a)$ should be computed over the possible entity space, which we approximate by \textit{taking all the named entities} from both evidence $E$ and the original claim $x^0$.
Thus, the distribution can be acquired by normalizing the perplexity of each entity predicted by T5 in the candidate entity space.

\paragraph{Replacement}
For replacement ($a=\mathtt{rep}$), we define $P_3(x^\prime \mid x_{-m}, \mathtt{rep})$ is either $P^\mathrm{ent}_3(x^\prime\mid x_{-m}, \mathtt{rep})$ or $P^\mathrm{tok}_3(x^\prime\mid x_{-m}, \mathtt{rep})$,
because T5 replaces the mask either with entity generation or token generation, depending on whether the mask is within an entity or not.

\paragraph{Insertion}
For \textit{insertion} ($a = \mathtt{ins}$), we reload the notation $x_{-m} = \{w_1, \cdots, w_{m}, \texttt{[MASK]}, w_{m+1}, \cdots, w_{|x|}\}$.
In other words, we insert a \texttt{[MASK]} after the position $m$ and keep the original tokens unchanged.
Since we do not know whether an entity or a token should be generated, we take a weighted distribution of the two, \ie, $x$, where $\alpha$ is set as 0.5 in practice.

\paragraph{Deletion}
For \textit{deletion} ($a = \mathtt{del}$), the returned text is $x^\prime = \{w_1, \cdots, w_{m-1}, w_{m+1}\cdots, w_{|x|}\}$, where the token/entity $w_m$ is deleted.
So, $P_3(x^\prime\mid x_{-m}, \mathtt{del}) = 1$, and its reverse is actually \textit{insertion}.

\subsection{Convergence Analysis}
\label{sec:converge}

We now give an analysis of the reversibility of the Markov chain, in response to the challenge mentioned at the beginning of $\mathsection$~\ref{sec:corrector}.
In our implementation, the state transition within entities or tokens can naturally be reversible due to the separation of two spaces.
To communicate two spaces, a delete-plus-insert action combination can be conducted.
For example, when replacing an entity with a token, we can first delete the entity and then insert a token, thanks to the insertion action that balances two spaces. 
As a result, we can theoretically ensure the correctness and convergence of our implementation.

\section{Experimental Setup}
\label{sec:experiments}

\subsection{Evaluation}

\paragraph{Dataset}

In this work, we evaluate our method using the dataset proposed by~\citet{thorne-vlachos-2021-evidence}, which is an evidence-based FEC task (referred to as \dataset for brevity).
\dataset is collected from the intermediate annotations in FEVER~\cite{thorne-etal-2018-fever}, a large and widely used fact verification dataset.
A claim in FEVER is verified to be \textsc{Supported} or \textsc{Refuted} by evidence, or labelled to be unverifiable, \ie, \textsc{NotEnoughInfo}.

\dataset takes the \textsc{Supported} and \textsc{Refuted} samples from FEVER.
Based on the \textsc{Refuted} samples, \dataset utilizes the mutated text spans that convert sentences from Wikipedia into \textsc{Refuted} claims.
Statistics of \dataset are reported in Table \ref{tab:dataset}.
Note that, \textsc{Supported} instances are also included in the test set to test the models for keeping a correct claim as it is.

\begin{table}[t]
\small
  \centering
    \begin{tabular}{lccc}
    \toprule
    \textbf{Label} & \textbf{\# Train} & \textbf{\# Valid} & \textbf{\# Test} \\
    \midrule
    \textsc{Supported} & 37,961 & 1,477 & 1,593 \\
    \textsc{Refuted} & 20,075 & 2,091 & 2,289 \\
    \bottomrule
    \end{tabular}%
  \caption{Statistics of \dataset \cite{thorne-vlachos-2021-evidence}, with data sample counts of each split and label.}
  \label{tab:dataset}%
\end{table}%

\paragraph{Metrics}

For automatic evaluation, we primarily use the SARI metric~\cite{xu-etal-2016-optimizing}, which measures the goodness (F1) of words that are \textit{added}, \textit{deleted} or \textit{kept} by the system from the source, compared with referenced ground truth.
SARI-\textit{final} is the average score of the three.
We also use ROUGE score (RG-2) \cite{lin2004rouge} for evaluating the information recalled from the reference.
\citet{thorne-vlachos-2021-evidence} show that, statistically, SARI scores and ROUGE score are \textit{highly} correlated with human judgments in this task, especially SARI, according to Pearson's $r$.

\subsection{Baseline Methods}

We adopt baselines of two settings: supervised and distantly supervised methods.
The distinction between them lies in whether direct supervision for FEC (the refuted claims and their corrections) is used for model training.

\paragraph{Supervised Baselines}

We train two models with \textsc{Refuted} claims and their corrections from the training set of \method, covering paradigms of autoregressive generation and editing:
\begin{itemize}
    \item T5~\cite{raffel-etal-2020-exploring}, a strong pre-trained baseline that excels in sequence-to-sequence tasks. 
    We fine-tune it by concatenating the original claim and evidence texts as input to generate a corrected claim.
    \item EdiT5~\cite{mallinson2022edit5}, a T5-based model that combines the best of non-autoregressive editing and autoregressive generation. 
    The editing operations in EdiT5 include tagging, reordering, and insertion.
\end{itemize}

\paragraph{Distantly Supervised Baselines}

Since resources for FEC are limited, previous work mostly focuses on distantly supervised methods to address this problem.
Specifically, two general types of strategies are used in previous work for FEC:
\begin{enumerate}
    \item \texttt{DS-1}: evidence-based mask token/span prediction trained on \textsc{Supported} data samples~\cite{Shah_Schuster_Barzilay_2020,thorne-vlachos-2021-evidence,Chen_Bao_Sun_Zhang_Chen_Zhou_Xiao_Li_2022}, a denoising training objective similar to that used in BERT~\cite{devlin-etal-2019-bert}, BART~\cite{lewis-etal-2020-bart} and T5; and
    \item \texttt{DS-2}: distant supervision from external discriminative tasks, such as natural language inference (NLI)~\cite{williams-etal-2018-broad} and fact verification~\cite{thorne-etal-2018-fever} (which is also a type of NLI). 
\end{enumerate}

Applying these strategies (either or both), our distantly supervised baselines include:
\begin{itemize}
    \item Masked Language Model (BERT,~\citealt{devlin-etal-2019-bert}), applying \texttt{DS-1} to greedily decode masked tokens.
    \item Dual encoder pointer network (2EncPtr for short)~\cite{Shah_Schuster_Barzilay_2020}, applying both \texttt{DS-1} and \texttt{DS-2} by querying a fact-checking model to predict the words to be masked, then conditionally replacing the masks with ground truth evidence.
    \item T5 Masker-Corrector (T5MC for short)~\cite{thorne-vlachos-2021-evidence}, which is the previous SOTA and thus our primary baseline. 
    T5MC follows a masker-corrector pipeline: \textsc{Supported} claims in \dataset are used to apply \texttt{DS-1} to train a corrector;
    during inference, masks are \textit{heuristically} decided, where text spans that do not appear in evidence texts will be masked.
    \item T5 Masker-Corrector with verification signals (T5MC-V for short)~\cite{thorne-vlachos-2021-evidence}, which is an extension to T5MC.
    T5MC-V follows~\cite{Shah_Schuster_Barzilay_2020} and uses an external fact verification model trained on FEVER to decide the tokens to be masked. 
    Thus, T5MC-V applies both \texttt{DS-1} and \texttt{DS-2}.
\end{itemize}
What if these one-pass FEC baselines also have $K$ candidates (same as the number iterations in \method) to be verified by FV?
To make a fairer comparison, we upgrade T5MC and T5MC-V by taking $K=20$ generated sequences of them, and use the same FV model to rank for the best one, denoted as ``Model+\textit{enumerate}''.

\subsection{Implementation Details}
\label{sec:implementation}

\paragraph{Evidence Retrieval}

Since the retrieval of evidence is not the primary focus of this work, we adopt an existing approach and keep the retrieved evidence to be the same for all systems.
Following previous work~\cite{thorne-vlachos-2021-evidence}, the retrieval module consists of two main steps: 
\begin{inparaenum}[\it 1)]
    \item given a claim, the contained entities are predicted using GENRE~\cite{cao2021autoregressive}, an autoregressive entity linking system, thus related pages are retrieved from Wikipedia; then
    \item top-$k$ ($k=2$) passages are selected based on the similarity between the claim and these pages using DPR~\cite{karpukhin-etal-2020-dense}, a dense retriever based on the BERT encoder.
\end{inparaenum}

\paragraph{Training Details}

We train the verifier (NLI) models and the proposal models based on the checkpoints hosted by HuggingFace~\cite{wolf-etal-2020-transformers}.
Most of the training scripts and parameters are by default of the \texttt{transformers} library from HuggingFace. Specifically,
\begin{inparaenum}[\it 1)]
    \item For the verifier (\ie, NLI models), the BERT (base version) and RoBERTa (large version) checkpoints are fine-tuned on the FEVER dataset. We trained the NLI model for 5 epochs at a learning rate of $1*10^{-5}$.
    \item For the proposal model, we fine-tune a T5 (base version) model with randomly masked \textsc{Supported} claims and the corresponding evidence in the \dataset dataset. We train the model for 5 epochs at the learning rate of $3*10^{-5}$.
\end{inparaenum}

\section{Results and Analysis}
\label{sec:results}

\subsection{Overall Performance}


\setlength\tabcolsep{3pt}
\begin{table}[t]
\small
  \centering
    \begin{tabular}{llccccc}
    \toprule
    \multirow{2}{*}{\textbf{Method}} & \multirow{2}{*}{\textbf{Verifier}} & \multicolumn{4}{c}{\textbf{SARI (\%)}} & \multirow{2}{*}{\textbf{RG-2}} \\
    \cmidrule{3-6} 
    & & \textbf{Keep} & \textbf{Delete} & \textbf{Add} & \textbf{Final} & \\
    \midrule
    \rowcolor[rgb]{ .949,  .953,  .961} \multicolumn{7}{c}{\textit{Fully Supervised}} \\
    T5-base  & - & 79.6 & 90.2 & 59.2 & 76.4 & 72.7 \\
    EdiT5-base  & - & \textbf{81.8} & \textbf{93.0} & \textbf{63.4} & \textbf{79.4} & \textbf{76.9} \\
    \midrule
    \rowcolor[rgb]{ .949,  .953,  .961} \multicolumn{7}{c}{\textit{Distantly Supervised}} \\
    MLM  & - & 56.1 & 52.9 & 7.8 & 38.9 & 42.7 \\
    2EncPtr   & BERT$_b$ & 34.5 & 48.1 & 1.7 & 28.1 & 34.8 \\
    T5MC   & -  & 65.2 & 62.7 & 15.5 & 47.8 & 50.3 \\
    \ \ +\textit{enumerate} & BERT$_b$ & 66.2 & \textbf{64.3} & 17.1 & 49.2 & 51.2\\
    T5MC-V   & BERT$_b$ & 61.1 & 54.3 & 19.4 & 44.9 & 42.0 \\
    \ \ +\textit{enumerate} & BERT$_b$ & 63.0 & 55.7 & 24.1 & 47.6 & 45.5\\
        \cdashlinelr{1-7}
    \multirow{2}{*}{\method}  &  BERT$_b$  & 66.0 & 60.1 & 34.8 & 53.6 & 57.7 \\
     & RoBERTa$_l$ & \textbf{67.1} & 61.9 & \textbf{36.0} & \textbf{55.0} & \textbf{59.1}  \\
    \bottomrule
    \end{tabular}
  \caption{The automatic evaluation results of \method compared with baselines. 
  Distantly supervised methods with verifiers apply the \texttt{DS-2} strategy during training.
  Verifier$_b$ and Verifier$_l$ denote the base and large version of the pre-trained language model that the Verifier uses.}
  \label{tab:main}
\end{table}

We present the main results in the \dataset dataset in Table~\ref{tab:main}. 
Generally, we have the following findings: 

Our method outperforms previous distantly supervised baselines by large margins, including previous state-of-the-art (T5MC,~\citealp{thorne-vlachos-2021-evidence}). 
\method achieves 53+ SARI final score and 57+ ROUGE-2 score.
However, supervised methods still beat distantly supervised ones, envisioning a large room for improvement.
Compared with end-to-end text generation (T5), the EdiT5 wins due to the editing nature of the FEC task.

When adopting \texttt{DS-2} (\ie, with a verifier), we notice that the baselines do not fully exploit the verification signals for FEC.
Especially, T5MC suffers from performance degradation when using a verifier to help select masks.
In contrast, \method with a better verifier (RoBERTa-large,~\citealp{liu2019roberta}) consistently achieves better results (+1.4 in SARI final score and +1.4 ROUGE-2 score).
This indicates the success of our deep integration of verification guidance for FEC.

We notice that \method achieves more significant improvement on SARI-Add, indicating the added words of \method make much more sense than the baselines.
This is because iterative correction offers a more delicate control over the generated text, whereas one-time correction cannot settle the number and conflict of proposed words for each mask.

When more corrections are enumerated and ranked, baselines results do improve 1.4 SARI points (T5MC) and 2.7 SARI points (T5MC-V), but not as good as \method. 
This shows the effectiveness of the iterative editing framework and deep integration of FV in \method.

\subsection{Analysis}

In response to the motivations of \method, we conduct detailed analyses to further understand \method and why it works.
In particular, the analysis is designed to find out whether and how the verification model and the iterative editing process benefit FEC.
For the rest of the analysis, we use a BERT-base verifier in \method, unless specified.

\paragraph{How does verification affect correction?}


\begin{table}[t]
  \small
  \centering
    \begin{tabular}{llccccc}
    \toprule
   \multirow{2}{*}{\textbf{Dataset}} & \multirow{2}{*}{\textbf{Verifier}} & \multirow{2}{*}{\textbf{Acc (\%)}} & \multicolumn{4}{c}{\textbf{SARI (\%)}} \\
    \cmidrule{4-7} 
    & & & \textbf{Keep} & \textbf{Delete} & \textbf{Add} & \textbf{Final} \\
     \midrule
    \multirow{2}{*}{MultiNLI} & BERT$_b$ & 84.6 & 63.9 & 57.0 & 30.1 & 50.3 \\
    & RoBERTa$_l$ & \textbf{90.2} & \textbf{65.1} & \textbf{58.9} & \textbf{32.2} & \textbf{52.0} \\
    \midrule
    \multirow{2}{*}{FEVER} & BERT$_b$ & 71.7 & 66.0 & 60.1  & 34.8 & 53.6 \\
     & RoBERTa$_l$ & \textbf{72.9} & \textbf{67.1} & \textbf{61.9} & \textbf{36.0} & \textbf{55.0} \\
    \bottomrule
    \end{tabular}%
  \caption{Ablation results of the verification model used in \method w.r.t. model sizes and trained datasets. We report accuracy of verifier models on test set of MultiNLI and FEVER, respectively.}
  \label{tab:verification}
\end{table}%

To answer the first question, we explore different sets of verifiers in \method, each of them alters the gradient-based position finding and the energy model in the correction process.
We conduct an ablation study of 
\begin{inparaenum}[\it 1)]
    \item the verification ability of the verifier models, instantiated with types and sizes of PLMs (BERT and RoBERTa); and 
    \item the datasets that the verifiers are trained upon, instantiated with MultiNLI~\cite{williams-etal-2018-broad} and FEVER~\cite{thorne-etal-2018-fever}.
\end{inparaenum}

We draw two findings based on the results in Table~\ref{tab:verification}: 
\begin{inparaenum}[\it 1)]
    \item Consistent with findings from Table~\ref{tab:main}, better verification performance contributes to better correction performance, which holds for both MultiNLI and FEVER verifier;
    \item Even using an out-of-domain NLI model as a verifier, \method still achieves better performance than baselines, demonstrating the effectiveness of our design.
\end{inparaenum}
Our results in this part also call back to the fact that verifiers can be practically utilized to constrain language models for solving complex tasks, \eg, math-word problems~\cite{cobbe2021training}.

\paragraph{Will more editing iterations help correction?}

\begin{figure}[t]
    \centering
\pgfplotsset{width=0.7\linewidth,height=0.65\linewidth,compat=1.17}
\footnotesize
\begin{tikzpicture}
\begin{axis}[
    xlabel={\# of Iterations},
    ylabel={SARI Final (\%)},
    xmin=-1, xmax=31,
    ymin=-1, ymax=61,
    xtick={0, 5, 10, 15, 30},
    ytick={0, 10, 20, 30, 40, 50, 60},
    legend pos=south east,
    ymajorgrids=true,
    xmajorgrids=true,
    grid style=dashed,
    x label style={at={(axis description cs:0.5,-0.125)},anchor=north},
    y label style={at={(axis description cs:-0.125,0.5)},anchor=south},
    legend style={nodes={scale=0.7, transform shape}}
]
\addplot[
    color=Blue,
    mark=o,
    mark size=2.5pt,
    ]
    coordinates {
    (0, 14.2)
    (5, 37.1)
    (10, 47.4)
    (15, 53.6)
    (30, 54.0)
    };
    \addlegendentry{\method}

\addplot[
    color=ForestGreen,
    mark=triangle,
    mark size=3pt,
    ]
    coordinates {
    (0, 14.2)
    (5, 21.9)
    (10, 35.3)
    (15, 40.8)
    (30, 39.4)
    };
    \addlegendentry{\method w/o energy function $\mathcal{E}_\mathtt{V}$}

\addplot[
    color=Maroon,
    mark=diamond,
    mark size=3pt,
    ]
    coordinates {
    (0, 14.2)
    (5, 17.5)
    (10, 22.1)
    (15, 27.5)
    (30, 37.2)
    };
    \addlegendentry{\method w/ random masking}

\end{axis}
\end{tikzpicture}

    \caption{Correction results at each iteration of \method and its variants. Results on iteration 0 come from input claim.}
    \label{fig:iteration}
\end{figure}
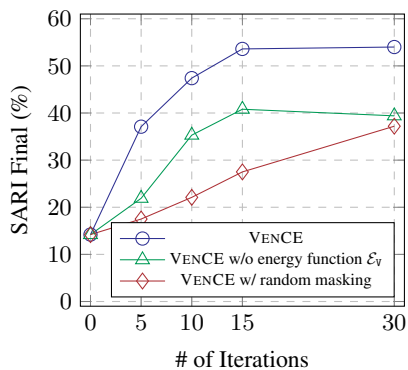
This question challenges the iterative editing design in \method.
To answer this question, we evaluate the results \method produces during each iteration ranging from 0 to 30, where Iter\#0 stands for the original claim.
Moreover, we also test two variants of \method, \ie, \method without verification model in the energy or without gradient-guided position finding.
Results are presented in Figure~\ref{fig:iteration}.
We find that 
\begin{inparaenum}[\it 1)]
    \item \method converges approximately at Iter\#15;
    \item when correcting without verification signals in the energy function $\mathcal{E}(x)$, \method suffers from a big performance drop ($\sim$13 points);
    \item the results for random masking shows a big acceleration of convergence when masking is guided with gradient-based sampling, where the verifier also plays an important role.    
\end{inparaenum}

\paragraph{How do different combinations of energy functions affect correction?}

\begin{table}[t]
    \centering
    \small
    \begin{tabular}{ccccccc}
    \toprule
        \textbf{Row} & {\textbf{Verif.}} & {\textbf{Hamm.}} & {\textbf{LM}} & \textbf{SARI-F} $\uparrow$  & \textbf{HammS.} $\downarrow$ & \textbf{BERTS.} $\uparrow$\\
    \midrule
        1 & \checkmark & & & 41.2 & 3.42 & 0.90 \\
        2 & & \checkmark & & 25.3 & 2.08 & 0.76 \\
        3 & & & \checkmark & 37.1 & \textbf{1.19} & 0.82 \\
        \cdashlinelr{1-7}
        4 & & \checkmark & \checkmark & 39.8 & 2.10 & 0.87 \\
        5 & \checkmark & & \checkmark & 49.7 & 2.36 & 0.92 \\
        6 & \checkmark & \checkmark & & 45.0 & 2.30 & 0.89 \\
        \cdashlinelr{1-7}
        7 & \checkmark & \checkmark & \checkmark & \textbf{53.6} & 2.21 & \textbf{0.93} \\
    \bottomrule
    \end{tabular}
    \caption{Correction results of \method with different combinations of energy functions.}
    \label{tab:energy}
\end{table}

One of the key modules in MH sampling and thus in \method is the stationary distribution $\pi(x)$ implemented with the probabilistic energy-based model $\mathcal{E}(x)$.
We enumerate the possible combinations of energy functions in \method and present the results in Table~\ref{tab:energy}.
SARI (final) measures correctness, HammingScore measures minimum modification, and BERTScore \cite{bert-score} measures fluency.
We find that
\begin{inparaenum}[\it 1)]
    \item $\mathcal{E}_\mathtt{V}$ for verified truthfulness contributes the most to the quality and correctness of the results, as seen in Row 1 \& 2 \& 3 and Row 4 \& 7;
    \item When only using $\mathcal{E}_\mathtt{LM}$, the claim is not edited much, thus HammingScore achieves the lowest (Row 3), whereas poor SARI score reveals its falsity.
    Also, SARI drops much when removing $\mathcal{E}_\mathtt{LM}$ (Row 6 \& 7), denoting the importance of fluency constraint during the correction.
    \item Interestingly, as seen in Row 5 \& 7, $\mathcal{E}_\mathtt{H}$ does not contributes to the Hamming distance score, but improves SARI for correctness (+3.9 points).
\end{inparaenum}
These findings indicate that all energy functions are indispensable for \method, and \method is successfully constrained to find the best-suited modification for a corrected claim.

\paragraph{What if we only edit tokens or entities?}

As in $\mathsection$~\ref{sec:masker}, \ref{sec:corrector}, we propose new edits from both token and entity spaces in order to accommodate for consecutive tokens within entities.
To show the effectiveness of this design, we conduct an ablation study where the correction happens \textit{only} on tokens or entities.
According to the results in Table~\ref{tab:tokenentity}, our design of generative proposal over two spaces greatly surpasses the token-only or entity-only counterparts, achieving a much higher score of over 30 points in SARI-Final.

\begin{table}[t]
    \centering
    \small
    \begin{tabular}{lcccc}
    \toprule
        \multirow{2}{*}{\textbf{Edit Choice}}  & \multicolumn{4}{c}{\textbf{SARI (\%)}} \\
    \cmidrule{2-5} 
     & \textbf{Keep} & \textbf{Delete} & \textbf{Add} & \textbf{Final} \\
    \midrule
        Entity-only  &  33.6  &  30.0  &  1.7  &  21.8 \\
        Token-only  &  34.5  &  28.6  &  9.1  &  24.7 \\
        Both  & \textbf{67.1}   &  \textbf{61.9}   & \textbf{36.0}  &    \textbf{55.0} \\
    \bottomrule
    \end{tabular}
    \caption{Correction results of \method, where the editing only happens on entities, tokens, or both.}
    \label{tab:tokenentity}
\end{table}

\paragraph{What if there is no error propagation from evidence retrieval?}

\begin{table}[t]
    \centering
    \small
    \begin{tabular}{llcccc}
    \toprule
        \multirow{2}{*}{\textbf{Method}} & \multirow{2}{*}{\textbf{Evidence}} & \multicolumn{4}{c}{\textbf{SARI (\%)}} \\
    \cmidrule{3-6} 
    & & \textbf{Keep} & \textbf{Delete} & \textbf{Add} & \textbf{Final} \\
    \midrule
        \multirow{2}{*}{T5MC} & \textit{retrieved} & 65.2 & \textbf{62.7} & 15.5 & 47.8 \\
         & \textit{gold} & 66.7 & 62.2 & 16.1 & 48.3 \\
        \cdashlinelr{1-6}
        \multirow{2}{*}{T5MC-V} & \textit{retrieved} & 61.1 & 54.3 & 19.4 & 44.9 \\
         & \textit{gold} & 61.8 & 62.2 & 10.2 & 44.7 \\
        \cdashlinelr{1-6}
        \multirow{2}{*}{\method} & \textit{retrieved} & 66.0 & 60.1 & \textbf{34.8} & 53.6 \\
         & \textit{gold} & \textbf{67.5} & 61.5 & 34.6 & \textbf{54.5} \\
    \bottomrule
    \end{tabular}
    \caption{Correction results of \method and baselines with retrieved (as in $\mathsection$~\ref{sec:implementation}) or ground truth evidence.}
    \label{tab:evidence}
\end{table}

Same to other retrieval-based tasks, the performance of correction is inevitably affected by the error propagation from the evidence retrieval module.
Since our focus in this paper is not evidence retrieval, we here present the ceiling performance when FEC systems correct claims with ground truth evidence.
Results in Table~\ref{tab:evidence} show that correcting with gold evidence does improve the results in most cases, but marginally.
We believe future upgrades of FEC methods should primarily concentrate on improving FEC itself.

\begin{table}[t]
    \small
    \centering
    \begin{tabular}{lccc}
    \toprule
        \textbf{Method} & \textbf{Grammatical} & \textbf{Supported} & \textbf{Corrected}  \\
    \midrule
        T5MC & 92.0 & 81.3 & 55.3 \\
        T5MC-V & 92.0 & 80.0 & 51.3 \\
        \method & \textbf{94.7} & \textbf{84.0} & \textbf{74.7} \\
    \bottomrule
    \end{tabular}
    \caption{Human evaluation results for corrected claims.}
    \label{tab:human}
\end{table}

\begin{table}[t]
    \small
    \centering
    \begin{tabularx}{\linewidth}{llc}
    \toprule
        \textbf{\# Iter.} & \textbf{Proposed Claims} & \textbf{Acc.} \\
    \midrule
        0 & One True Thing is a German film. & - \\
        \cdashlinelr{1-3}
        1 & One True Thing is a film. & \textcolor{red}{\cmark} \\
        2 & One True Thing is film. & \xmark \\
        3 & One True Thing is a drama film. & \textcolor{red}{\cmark} \\
        4 & One True Thing is drama film. & \xmark \\
        5 & One True Thing is American drama film. & \textcolor{red}{\cmark} \\
        6 & One True Thing is an American drama film. & \textcolor{red}{\cmark} \\
        7 & One True Thing is an American film. & \textcolor{red}{\cmark} \\
        8 & One True Thing is an American drama. & \xmark \\
        9 & One True Thing is an 1998 American film. & \xmark \\
        10 & One True Thing is an American. & \xmark \\
        11 & \textcolor{red}{\textbf{One True Thing is an American drama film.}} & \textcolor{red}{\cmark} \\
        12 & One True Thing is an American drama. & \xmark \\
        13 & One True Thing is an American film.  & \xmark \\
        14 & One True Thing is a American drama film. & \xmark \\
        15 & One True Thing is an American film. & \xmark \\
        \cdashlinelr{1-3}
        - & One True Thing is an American film. & Gold \\
    \bottomrule
    \end{tabularx}
        \caption{The proposed claim of \method and its acceptance (\textbf{Acc.}) decisions for every iteration (\textbf{\# Iter.}). Text in bolded red is the output corrected claim.}
    \label{tab:visualize_iteration}
\end{table}

\paragraph{Human Evaluation}

Finally, we manually evaluate the FEC performance of \method and baselines, even though SARI and ROUGE scores are reported to be \textit{highly} correlated with human judgment \cite{thorne-vlachos-2021-evidence}.
We sample 100 cases and ask three students (undergraduate and graduate) to answer three Boolean questions for each instance 
\begin{inparaenum}[\it 1)]
    \item \textit{is it grammatically correct?} 
    \item \textit{is it supported by evidence? }
    \item \textit{are the original errors corrected?}
\end{inparaenum}
We take the average of the results of three annotators.
After a training session, in the annotation session, they reach an inter-annotator agreement at Fleiss's $\kappa=0.56$.
The results are shown in Table~\ref{tab:human}.
\method consistently outperforms baseline systems, especially in the third dimension (+20 points), \ie, ``are the errors corrected?''.
Also, results by \method are better supported by the evidence and more grammatically accurate.
These findings suggest that both our method and baselines can generate (or copy) reasonable results from evidence to the output, but \method better concentrates on the correction of given errors.

\paragraph{Editing History}
\label{sec:casestudy}

We present the history the proposed corrected claim of each iteration and its acceptance decision in Table~\ref{tab:visualize_iteration}.
The original claim gradually moves towards being supported by evidence after 15 rounds of iterative editing.
Most accepted proposals happen in the early iterations because the correction of apparent errors is more easily accepted.
Since the correctness of the claim has almost been reached in iteration 18, the gradient-based distribution is not as sharp as that in the previous iterations.
Even though such a position is carelessly sampled, the correction proposal will not be accepted.

\section{Conclusion}
\label{sec:conclusion}
In this paper, we target the factual error correction problem and propose a novel approach (\method) for this task.
\method is built on the idea of iterative editing and fact verification constrained correction, which further advances the new state-of-the-art of distantly supervised approach for FEC when high-quality training data is lacking.
Further analysis also proves that the iterative editing converges well, and truthfulness signals from verification serves as an indispensable component for \method in FEC.
However, our work is still limited by the external fact verification model that we used, which is not by design able to tell a claim that is \textit{more} supported than another.
To tackle this, future work could design 
better verification models that know \textit{degrees} of factual errors.
Another limitation is that \dataset is still somehow too small and simple to reflect real-world scenarios of factual errors in texts, where we call for more comprehensive datasets for this task of increasing importance.

\section*{Acknowledgement}
We thank the anonymous reviewers for their valuable comments and suggestions.
This work was supported by Science and Technology Commission of Shanghai Municipality Grant (No. 22511105902) and Shanghai Municipal Science and Technology Major Project (No.2021SHZDZX0103).
Yanghua Xiao is also a member of Research Group of Computational and AI Communication at Institute for Global Communications and Integrated Media, Fudan University.

\bibliography{aaai23}

\end{document}